\PassOptionsToPackage{main=english,russian}{babel}
\PassOptionsToPackage{T2A}{fontenc}
\documentclass[onecolumn]{misraj}

\usepackage{lipsum}
\usepackage{colortbl}
\usepackage{tabulary}
\usepackage{longtable}
\usepackage{array}
\usepackage[section]{placeins} 
\usepackage{tablefootnote}
\usepackage{tcolorbox}
\usepackage{wrapfig}
\usepackage{adjustbox}
\usepackage{float}
\usepackage{caption}

\usepackage{pifont}
\usepackage{multirow}
\usepackage{tabularx}
\usepackage{arydshln}
\usepackage{xspace}
\usepackage{makecell} 

\usepackage[framemethod=TikZ]{mdframed}
\usepackage{multicol}

\usepackage[utf8]{inputenc}
\usepackage[T1]{fontenc}
\usepackage[english]{babel}

\definecolor{lightblue}{RGB}{211, 227, 252} 
\definecolor{bgblue}{RGB}{247, 250, 255} 

\newcommand*\colourcheck[1]{%
  \expandafter\newcommand\csname #1check\endcsname{\textcolor{#1}{\ding{52}}}%
}
\newcommand*\colourcross[1]{%
  \expandafter\newcommand\csname #1cross\endcsname{\textcolor{#1}{\ding{55}}}%
}
\DeclareSymbolFont{extraup}{U}{zavm}{m}{n}
\DeclareMathSymbol{\vardiamond}{\mathalpha}{extraup}{87}
\colourcheck{blue}
\colourcheck{green}
\colourcheck{red}

\colourcross{blue}
\colourcross{green}
\colourcross{red}

\usepackage{nicematrix}
\usepackage{comment}
\usepackage{amssymb}
\usepackage[bottom]{footmisc}
\usepackage{svg}
\usepackage{spverbatim}

\usepackage{algorithm}
\usepackage{algpseudocode}
\usetikzlibrary{arrows.meta,positioning,fit,calc,backgrounds}

\RequirePackage[hidelinks,colorlinks=true,linkcolor=blue,citecolor=blue]{hyperref}

\setcitestyle{number,square}
\definecolor{deeppurple}{HTML}{9e02f7}
\definecolor{forestgreen}{HTML}{2e7d43}

\hypersetup{urlcolor=deeppurple}

\newcommand{\method}{Fixer-Guided Localization\xspace}
\newcommand{\methodshort}{FGL\xspace}
\newcommand{\ystar}{y^{\star}}
\newcommand{\yplus}{y^{\star}_{+}}
\newcommand{\ctx}{\mathrm{ctx}}

\newcommand{\stage}[1]{\texttt{#1}}

\title{Let the Fixer Decide: Training a Code Localizer from the Fixer's Likelihood}

\multiauthors
\multiauthors
\author{
    name={Khalil Hennara},
    email={hennara@misraj.ai}
}
\author{
    name={Muhammad Hreden},  
    email={hreden@gmail.com}
}
\author{
    name={Zeina Aldallal},
    email={aldallal@misraj.ai}
}
\author{
    name={Sara Chrouf},
    email={sara.chrouf@misraj.ai}
}
\author{
    name={Safwan AlModhayan},
    email={safwan@misraj.ai}
}

\affiliations{
 \includegraphics[width=2cm]{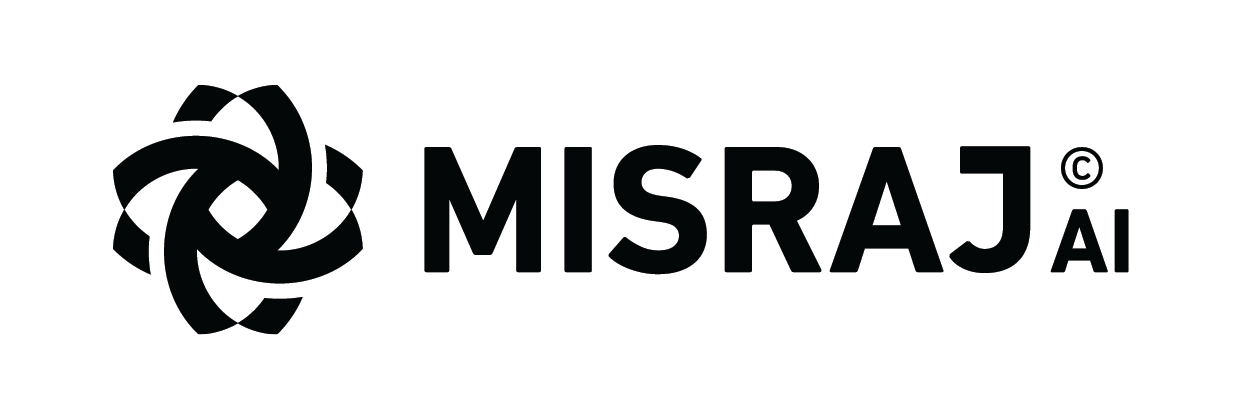}\\
     Khobar, Saudi Arabia \\
     \{hennara,hreden,aldallal,chrouf,safwan\}@misraj.ai \\
}
\date{\today}
\date{\today}
\abstract{
Code localizers are usually trained to predict the files a reference patch edits, but the code a fixer needs to read is not the same as the code it changes. We propose \emph{\method} (\methodshort): the localizer proposes several localizations, a frozen fixer scores each by how likely it makes a verified patch, and the localizer is trained on these scores with weighted supervised fine-tuning, without running tests and without reinforcement learning. We build the full pipeline on Gemma~4 E4B for the \textit{Kaggle Gemma-4 Developer Agent competition} \textsuperscript{*}, from 1,629 test-verified Kimi~K3 trajectories. The \methodshort target helps the fixer find the bug as often as the teacher's localization while using 27.5\% less code. On the development set, one EM round moves the agent in the expected direction compared with a matched control: more tasks reach a patch (53 vs.\ 46), fewer tasks get stuck reading files that do not exist (43 vs.\ 54), and more tasks are solved (8 vs.\ 6). Training also changes how the agent works: trained agents explore about three times more than the base model and write far fewer wrong patches, and the single agent solves up to 11 of 78 tasks, against 8 for the base model. An analysis of 13 evaluation runs identifies what limits a 4B agent under an 8-minute budget (time, repetition loops and edit quality) and gives practical lessons for training small coding agents. All data, code and adapters are publicly available.
}

\begin{document}

\renewcommand{\thefootnote}{\fnsymbol{footnote}}
\footnotetext[1]{\url{https://www.kaggle.com/competitions/gemma-4-developer-agent-paper}}
\section{Introduction}
\label{sec:introduction}

Language-model agents now fix many real GitHub issues on benchmarks such as SWE-bench~\citep{jimenez2024swebench,openai2024verified}. Some explore and edit the repository freely with shell tools~\citep{yang2024sweagent,wang2025openhands}; others first find the relevant code and then write a patch~\citep{xia2024agentless}. When open models are trained for the second design, the localizer learns to predict the locations that the reference patch edits~\citep{xie2025swefixer,tang2025copatcher}.

That label is a weak stand-in for what the fixer needs. A fix often depends on code it does not change, such as a caller, a base class or a test, and a localizer with about 80\% file accuracy can still give only 38\% resolved issues~\citep{mamun2026blagent}. In retrieval-augmented language models the answer is well known: train the retriever on what helps the reader~\citep{lewis2020rag,sachan2021emdr2,izacard2023atlas,shi2024replug}. For issue resolution, learning from outcomes has meant reinforcement learning with test execution~\citep{wei2025swerl,ma2025sorft}, which is expensive. We ask whether a localizer can learn from the fixer with supervised fine-tuning alone.

\method (\methodshort) treats the localization as a hidden variable. The localizer proposes candidates, a frozen fixer scores each by the log-probability it gives to a verified patch, and the localizer is fine-tuned on the candidates weighted by these scores. This is expectation-maximization (EM)~\citep{dempster1977em}; every update is ordinary weighted fine-tuning, and the only ``environment'' is a forward pass of the fixer.

We built the method for the Kaggle Gemma~4 Developer Agent competition~\citep{gemma-4-developer-agent-paper}, which fixes the model, the serving stack and the tools. With a limited budget we trained only the smallest model, Gemma~4 E4B. The result is mixed, and we report it as it is: the method works offline and moves the agent's behaviour in the right direction, but on 78 development tasks no trained system is significantly better than the untrained model. We study 13 evaluation runs to explain why.

\paragraph*{Contributions.}
\begin{itemize}[leftmargin=*]
    \item \textbf{Method.} A localizer trained from the fixer's likelihood of a verified patch, with weighted supervised fine-tuning only \footnote[2]{\url{https://github.com/misraj-ai/locfix}}(\Cref{sec:method}).
    \item \textbf{Data.} A pipeline that generates and verifies teacher trajectories inside the competition's own agent loop and sandbox \footnote[3]{\url{https://github.com/khalil-Hennara/codeagent-validate}}, a pipeline that converts them into two-agent training data without inventing any tool output, and the released set of 1,629 verified trajectories (\Cref{sec:data}).
    \item \textbf{Evaluation.} Eight systems on Gemma~4 E4B, compared with paired tests on the same tasks, including a matched control for the EM round (\Cref{sec:results}).
    \item \textbf{Analysis.} What limits a small agent under a strict time budget, with practical lessons for others (\Cref{sec:failures,sec:discussion}).
\end{itemize}

\section{Method}
\label{sec:method}

\subsection{Two agents}
\begin{figure}[tbp]
\centering
\begin{tikzpicture}[
    font=\small,
    box/.style={draw=black!55, rounded corners=3pt, align=center, inner sep=6pt, fill=white},
    agent/.style={box, draw=blue!60!black, line width=0.9pt, fill=blue!4},
    arr/.style={-{Stealth[length=2.2mm]}, draw=black!70, line width=0.6pt},
    lbl/.style={font=\footnotesize, text=black!75, align=center}
]
\node[box, text width=2.3cm] (issue) at (0,0) {Issue $x$\\ repository $R$\\ at base commit};
\node[agent, text width=6.0cm] (fixer) at (5.4,0) {\textbf{Fixer}: main agent, adapter $\theta$\\[3pt]
  \footnotesize tools: \texttt{run\_command}, \texttt{read\_file},\\
  \footnotesize \texttt{edit\_file}, \texttt{write\_file}, \texttt{get\_status},\\
  \footnotesize \texttt{submit\_patch}, and \texttt{localizer}\\[2pt]
  \footnotesize reads $z$, requests more, edits, tests};
\node[box, text width=2.3cm] (patch) at (12.0,0) {Patch $y$\\ graded by\\ hidden tests};
\node[agent, text width=6.0cm] (loc) at (5.4,-4.3) {\textbf{Localizer}: sub-agent, adapter $\phi$\\[3pt]
  \footnotesize tools: \texttt{run\_command}, \texttt{read\_file}\\
  \footnotesize fresh session, at most 15 calls};
\node[box, text width=2.3cm, draw=black!35] (repo) at (12.0,-4.3) {Repository in a\\ fresh container\\ (no network)};
\draw[arr] (issue) -- (fixer);
\draw[arr] (fixer) -- node[lbl, above] {submit} (patch);
\draw[arr] ([xshift=-1.4cm]fixer.south) -- node[lbl, left, xshift=-3pt] {call\\ \texttt{localizer(request)}} ([xshift=-1.4cm]loc.north);
\draw[arr] ([xshift=1.4cm]loc.north) -- node[lbl, right, xshift=3pt] {\texttt{\{"result": z\}}:\\ $\le 8$ locations,\\ $\approx$6K tokens of code} ([xshift=1.4cm]fixer.south);
\draw[arr, <->] (loc) -- node[lbl, above] {explore} (repo);
\end{tikzpicture}
\caption{The two-agent structure. One frozen base model carries two LoRA adapters. The fixer calls the localizer as a tool, receives a list of locations $z$, reads them, may request more code, then edits, tests and submits. Both agents share the task's tool-call budget.}
\label{fig:agents}
\end{figure}
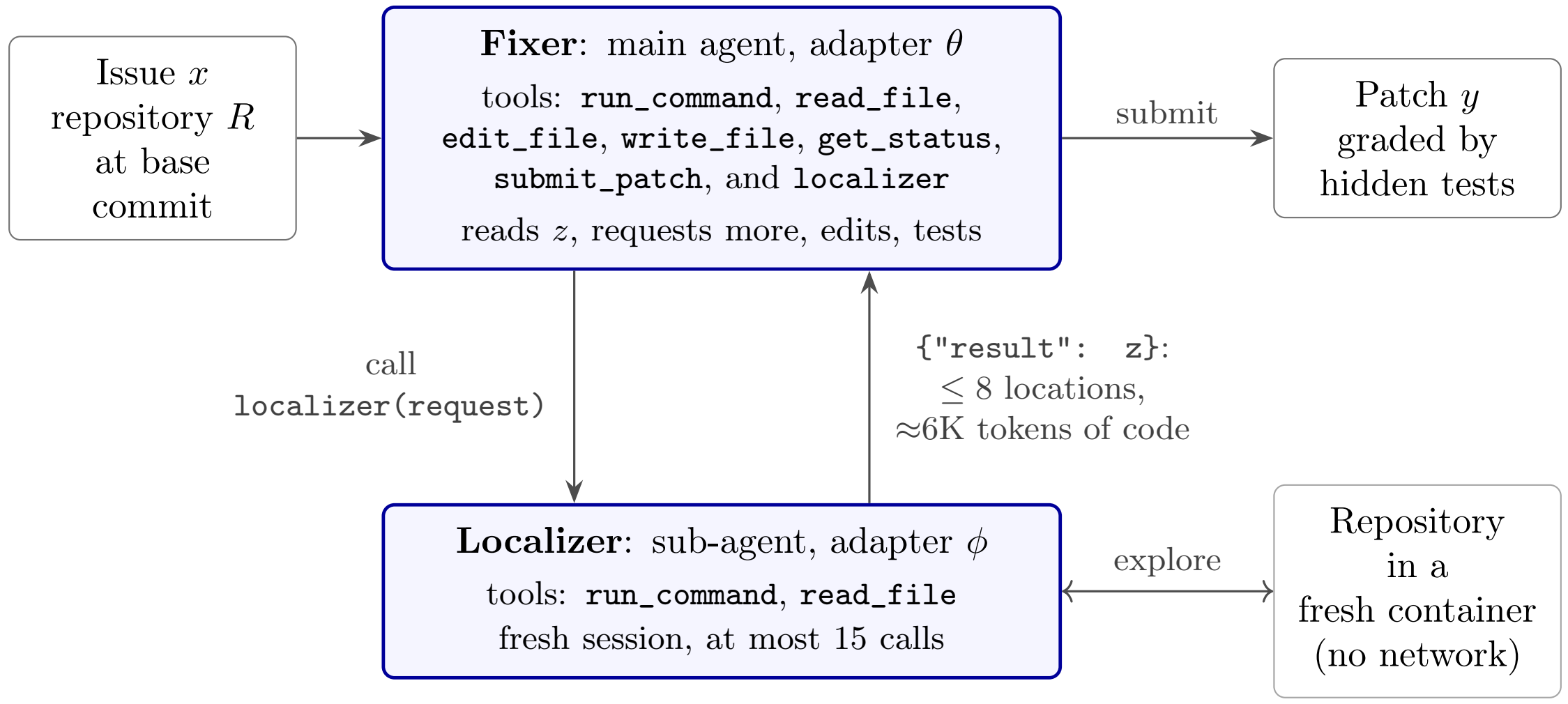

One frozen base model carries two LoRA adapters~\citep{hu2022lora} (\Cref{fig:agents}). The \emph{fixer} is the main agent. It calls the \emph{localizer} as a tool; the localizer explores the repository in a fresh session and answers with a list of at most 8 code locations (path, symbol, line range, and whether the location is to be edited or only read). The fixer reads these locations, may read more code (a \emph{request}), then edits, tests and submits.

\subsection{Localization as a hidden variable}
For a task we have the issue $x$, the repository $R$ and the teacher's patch $\ystar$, which passed the task's tests. The localizer gives $p_\phi(z \mid x, R)$ over localizations $z$, and the fixer gives $p_\theta(y \mid x, \ctx(z))$, where $\ctx(z)$ is the code that $z$ points to. We never see the ``right'' $z$, so we maximize the likelihood of the verified patch over all of them:
\begin{equation}
\label{eq:marginal}
\log p(\ystar \mid x, R) \;=\; \log \sum_{z} p_\phi(z \mid x, R)\; p_\theta\big(\ystar \mid x, \ctx(z)\big).
\end{equation}
A localization is judged only by how much it helps the fixer write the verified patch. The gradient for the localizer is an average under the posterior
\begin{equation}
\label{eq:posterior}
q(z) \;\propto\; p_\phi(z \mid x, R)\; p_\theta\big(\ystar \mid x, \ctx(z)\big),
\end{equation}
so training should raise the probability of localizations that make the patch likely and lower the others.

\subsection{Scores and weights}
We draw $K$ candidates $z_1,\dots,z_K$ from the localizer itself. With the localizer as the proposal, the posterior weights become a softmax over the fixer's scores. We score each candidate by
\begin{equation}
\label{eq:score}
s_k \;=\; \frac{1}{|\yplus|}\sum_{t \in \yplus} \log p_\theta\big(\ystar_t \mid \ystar_{<t},\, x,\, \ctx(z_k)\big)\;-\;\lambda_{\mathrm{ctx}}\,\mathrm{tok}(z_k)\;-\;\lambda_{\mathrm{req}}\,|a_k|,
\end{equation}
and set $w_k = \mathrm{softmax}_k(s_k/\tau)$. Three choices make this work in practice:
\begin{itemize}[leftmargin=*]
    \item \textbf{Only the new lines.} $\yplus$ is the set of tokens on the patch's added lines. Removed and context lines can be copied from $\ctx(z)$, so scoring them mostly rewards ``the edited file is shown''.
    \item \textbf{A cost for size and requests.} Without a cost, the best localization is ``show everything''. $\mathrm{tok}(z_k)$ is the size of the localized code and $|a_k|$ the number of extra reads the fixer needs.
    \item \textbf{A target effective sample size.} Raw score gaps make the weights one-hot. We choose $\tau$ per task so that the effective sample size is $n^\star$: $n^\star=2$ (\emph{soft} EM) keeps about two candidates; $n^\star=1$ (\emph{hard} EM) keeps the best one.
\end{itemize}
Both penalties are set to $0.25\,\sigma$, where $\sigma=0.068$ nats per token is the median spread of scores within a task.

\subsection{Training}
\begin{figure}[tbp]
\centering
\includegraphics[width=0.9\linewidth]{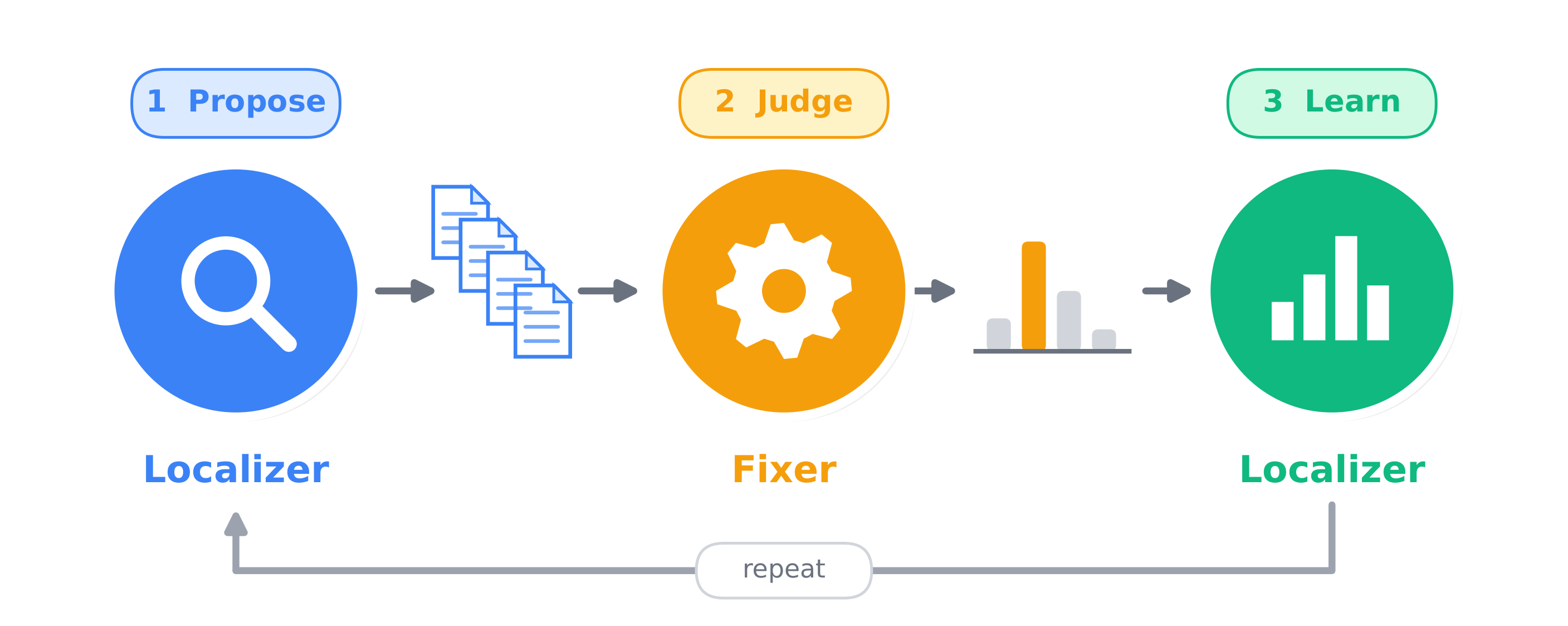}
\caption{One round of \methodshort. The localizer proposes $K$ candidate localizations; a frozen fixer scores each by its likelihood of the verified patch $\ystar$; the normalized scores weight ordinary fine-tuning of the localizer. No tests run and no reinforcement learning is used.}
\label{fig:hero}
\end{figure}
The localizer loss is a weighted negative log-likelihood of the candidates plus an anchor on the teacher's own localization $z^{\mathrm{teacher}}$ with weight $\beta=0.5$:
\begin{equation}
\label{eq:lloc}
\mathcal{L}_{\mathrm{loc}}(\phi) = -\sum_i \sum_{k} w_{ik}\,\log p_\phi(z_{ik} \mid x_i, R_i) \;-\; \beta \sum_i \log p_\phi\big(z^{\mathrm{teacher}}_i \mid x_i, R_i\big).
\end{equation}
\paragraph*{1. Two folds.} We split the training tasks into two halves, fold $A$ and fold $B$, and train one localizer and one fixer on each half. These fold models are used only to propose and score candidates; they are never deployed. The reason is memorization. A fixer that was trained on a task's patch can reproduce that patch from almost any context, so every candidate would get a high score and the weights would carry no information. In the same way, a localizer trained on a task mostly repeats the teacher's answer, so there would be little to choose from. Using the other fold's models avoids both problems: every task is handled only by models that never saw it.

\paragraph*{2. Candidates.} For each task, the localizer of the other fold proposes $K=8$ localizations (temperature 0.8). We add the teacher's localization $z^{\mathrm{teacher}}$ and a few simple variants (only the edited code, only the viewed code, the teacher's answer minus one item). After removing duplicates, a task has 7.1 candidates on average.

\paragraph*{3. Scores.} The fixer of the other fold reads each candidate and gives the score $s_k$ of \cref{eq:score}: the average log-probability of the patch's new lines, minus the costs for size and requests. This is one forward pass per candidate, with no gradient and no test run.

\paragraph*{4. Weights.} The scores become weights $w_k$ with the per-task temperature of the previous section. For each task we keep the three candidates with the highest weight and add the teacher's localization as an anchor with weight $\beta=0.5$, so the localizer does not drift too far from it. The weights of a task sum to one.

\begin{figure}[tbp]
\centering
\includegraphics[width=0.85\linewidth]{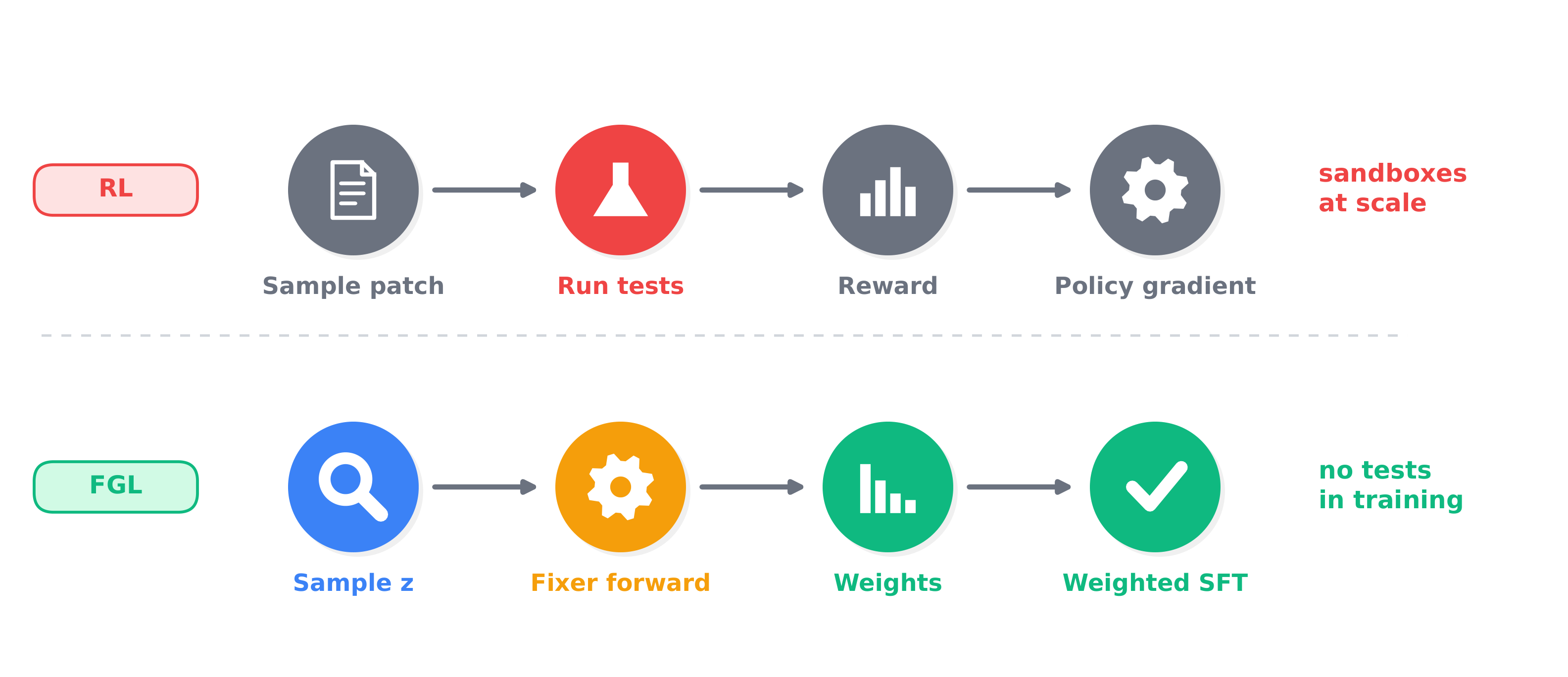}
\caption{Where the training signal comes from. \emph{Top:} reinforcement learning runs every sampled patch against tests in a sandbox. \emph{Bottom:} \methodshort scores each localization with one forward pass of the frozen fixer over the verified patch and trains with weighted fine-tuning.}
\label{fig:rl_vs_fgl}
\end{figure}

\paragraph*{5. Fine-tuning.} The deployed localizer, trained on all tasks, is fine-tuned on these weighted candidates with the loss of \cref{eq:lloc}, starting from its previous version at half the learning rate. This is ordinary weighted supervised fine-tuning: there are no online rollouts and no test runs (\Cref{fig:rl_vs_fgl}).

The full method also retrains the fixer on the new localizations in alternation with the localizer. In this paper we run one round and update only the localizer; every two-agent system uses the same fixer.

\section{Data}
\label{sec:data}

Data was the most expensive part of the project, (\Cref{tab:cost}).

\begin{table}[tbp]
\centering
\small
\caption{Cost of the project. All training and Stage~3 work (candidate generation, scoring, signal test) ran on A100 80\,GB GPUs; evaluation ran on Kaggle.}
\label{tab:cost}
\begin{tabular}{lr}
\toprule
Item & Cost \\
\midrule
Teacher trajectories (Kimi~K3, 3,068 attempts) & \$775 \\
Recall cleaning (two passes) & $\approx$\$100 \\
\midrule
Data, total & $\approx$\$875 \\
GPU time: all training, candidate generation, scoring and signal test & $\approx$130 A100 hours \\
\bottomrule
\end{tabular}
\end{table}

\begin{figure}[tbp]
\centering
\includegraphics[width=0.8\linewidth]{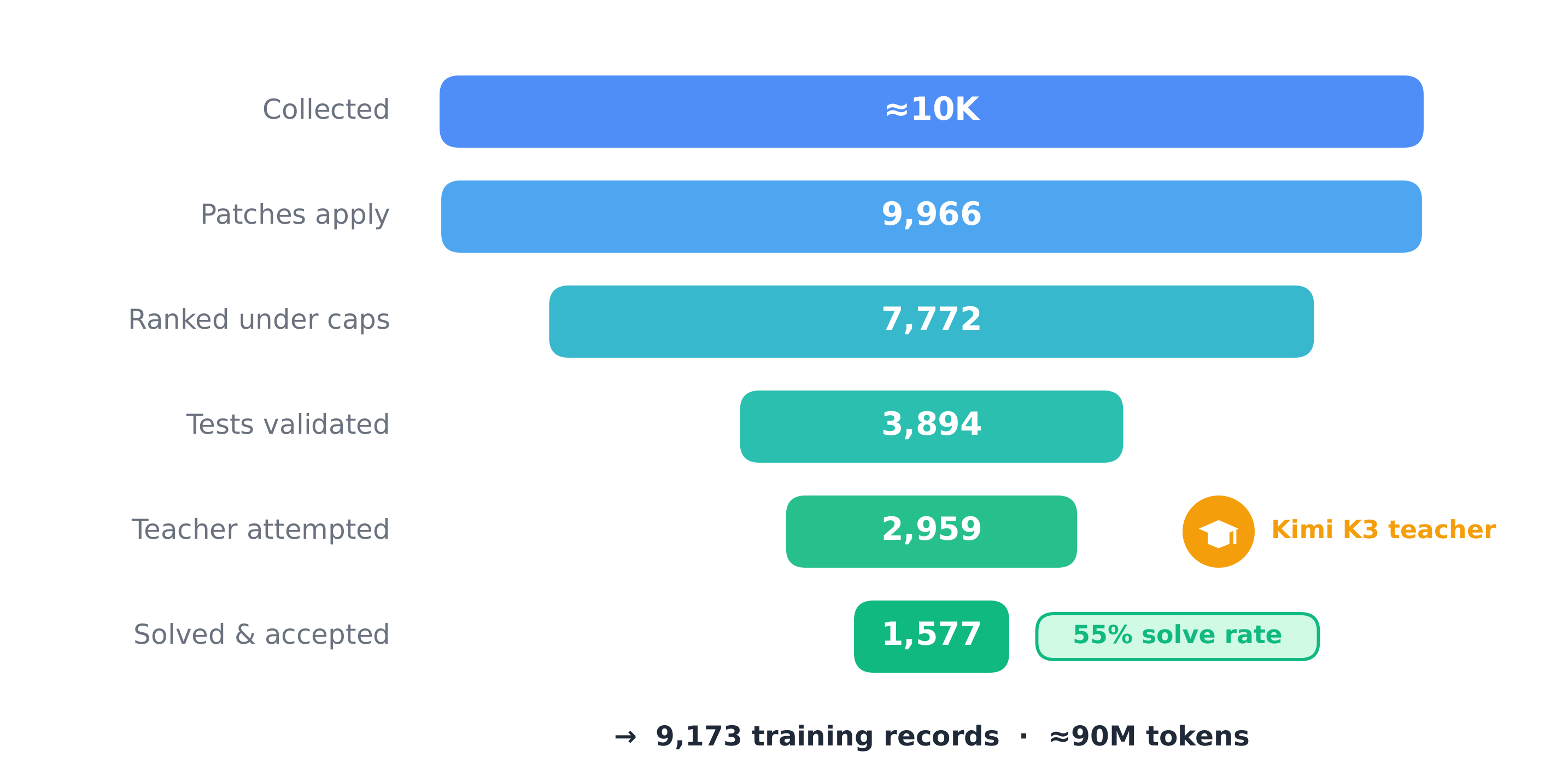}
\caption{From collected tasks to training records (not to scale; full counts in \cref{tab:funnel}).}
\label{fig:funnel}
\end{figure}

\paragraph*{Tasks.} We collected about 10K tasks from SWE-bench~\citep{jimenez2024swebench}, SWE-Gym~\citep{pan2025swegym}, R2E-Gym~\citep{jain2025r2egym} and SWE-smith~\citep{yang2025swesmith}. Every patch was checked to apply at its base commit, and a run without any model checked that the target tests fail before the reference fix and pass after it: 3,894 of the first 4,500 ranked tasks passed. Tasks from the four repositories of the competition's development set are excluded, so there is no overlap with the evaluation. \Cref{fig:funnel} and \cref{tab:funnel} in \cref{app:funnel} give every count.

\paragraph*{Teacher trajectories.} The teacher, Kimi~K3~\citep{kimik3}, ran inside the competition's own agent loop on each task's container, with 8 minutes per attempt and no network. Git history that could reveal the fix was removed first. A fresh container then ran the task's tests on the teacher's patch; a trajectory counts as \emph{solved} only if they pass. The code of generation and validation is released as \texttt{codeagent-validate}\footnote[2]{\url{https://github.com/khalil-Hennara/codeagent-validate}}. Of 3,068 attempts on 2,959 tasks, 1,629 tasks were solved (55\%): 564 from SWE-Gym, 492 from SWE-bench, 307 from SWE-smith and 266 from R2E-Gym. 1,577 of them were kept under per-source and per-repository caps. Only verified trajectories are used for training. A solved trajectory has a median of 15 tool calls (90th percentile 33) and 13.8K tokens.

\paragraph*{Recall cleaning.} The teacher often ``remembered'' the real upstream fix: it cited pull requests and commit hashes, some invented, and sometimes wrote a fix before reading the code. A student cannot do this at test time, and training on such text teaches it to state facts it never saw. We asked the teacher not to do this, and then ran a cleaning pass that found 3,233 recall passages in 1,116 trajectories and rewrote only those reasoning passages. Tool calls, tool outputs and patches were never changed.

\begin{figure}[tbp]
\centering
\includegraphics[width=0.9\linewidth]{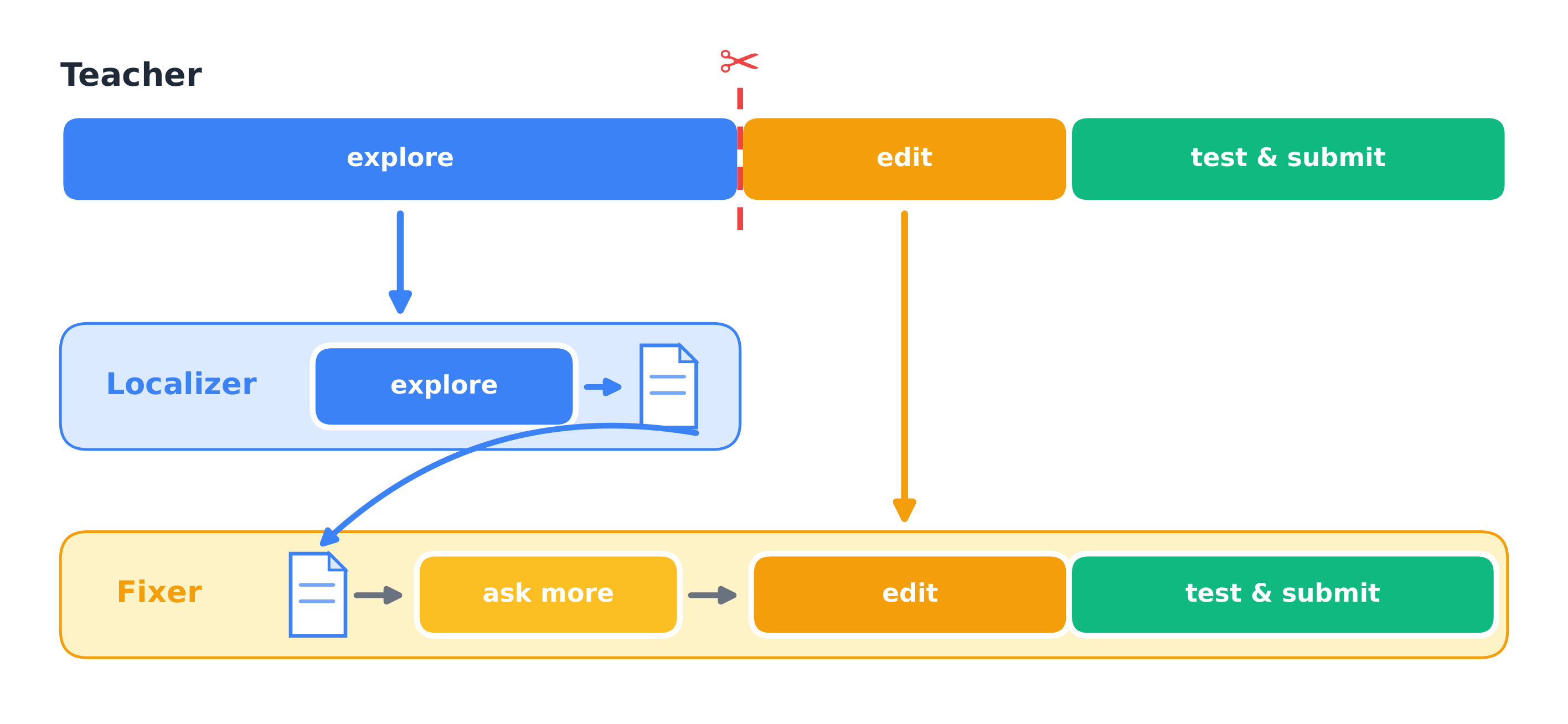}
\caption{One teacher trajectory becomes two records. The trajectory is cut at the first edit. The exploration becomes the localizer record ending in $z$; the fixer record reads $z$, may request missing code using only outputs the teacher saw, then continues with the teacher's edits unchanged.}
\label{fig:conversion}
\end{figure}

\paragraph*{From one agent to two.} The teacher works as a single agent. We cut each trajectory at its first edit (\Cref{fig:conversion}). The exploration before the cut becomes a localizer record that ends with an answer $z$. The fixer record calls the localizer, reads every location of $z$, may request code that $z$ misses, and then continues with the teacher's own edits and submission, unchanged. Every requested read is answered with code the teacher actually saw, checked byte for byte; no tool output is invented. The few steps we add (calling the localizer, reading $z$, asking for missing code) use short fixed sentences, for example ``I also need $X$, which the code I have read so far refers to. Let me read it.'' We used fixed text on purpose: a language model writing these lines could bring recalled knowledge back.

\paragraph*{Exact training text.} A trajectory does not contain the exact request the model receives: tool schemas, system-prompt additions and message structure come from the agent framework. We therefore replay every trajectory through the competition's own agent stack with a scripted model that returns the teacher's turns, and keep the requests it sends. Our token ids match vLLM's own rendering on all 141 test prompts. \Cref{tab:dataset} lists the resulting 9,173 records (about 90M tokens).

\begin{table}[tbp]
\centering
\small
\caption{Training records (Gemma~4 tokenizer). ``Trained'' is the share of tokens with a loss.}
\label{tab:dataset}
\resizebox{\linewidth}{!}{%
\begin{tabular}{llrrcrr}
\toprule
Records & Role & Records & Tasks & Tokens p50 / p90 / max & Total tokens & Trained \\
\midrule
\stage{s1\_agent} & single agent & 1,386 & 1,384 & 13K / 22K / 32K & 19.2M & 23.0\% \\
\stage{s2\_fix\_gold} & fixer, teacher's $z$ & 1,294 & 1,292 & 10K / 17K / 31K & 15.0M & 20.4\% \\
\stage{s2\_fix\_noisy} & fixer, imperfect $z$ + requests & 2,729 & 1,205 & 11K / 19K / 32K & 33.3M & 20.8\% \\
\stage{s2\_loc\_edit} & localizer, edited code & 1,286 & 1,286 & 5K / 11K / 20K & 8.0M & 14.6\% \\
\stage{s2\_loc\_view} & localizer, viewed code & 1,175 & 1,175 & 6K / 11K / 20K & 7.5M & 16.9\% \\
\stage{s3\_loc\_r0} & localizer, edited + viewed & 1,303 & 1,303 & 6K / 11K / 20K & 8.4M & 17.1\% \\
\bottomrule
\end{tabular}}
\end{table}

\section{Experimental Setup}
\label{sec:setup}

\paragraph*{Model and training.} All adapters \footnote[5]{\url{https://huggingface.co/Hennara/locfix-e4b-runs_p}} are trained on Gemma~4 E4B~\citep{gemma4}. The competition serves an int4 checkpoint, so we train bf16 LoRA (rank 64) on a dequantized copy and serve the adapter unmerged on the int4 model, as the scorer does; log-probabilities differ by 0.03 nats per token between training and serving. Each model is trained for about 250 optimizer steps of 16 sequences of up to 32,768 tokens, at learning rate $2\times10^{-4}$ (half for warm starts). Settings are in \cref{app:settings}.

\paragraph*{Systems.}
\begin{itemize}[leftmargin=*]
    \item \textbf{Base}: the untrained model as a single agent.
    \item \textbf{S1}: one adapter trained on the full teacher trajectories.
    \item \textbf{S2a / S2a-view}: localizer and fixer trained separately; the localizer learns the edited code (S2a) or the viewed code (S2a-view).
    \item \textbf{S2b}: as S2a, with the fixer also trained on imperfect localizations and requests. This is the fixer for every Stage~3 system.
    \item \textbf{S3r0}: localizer trained on the teacher's localization (edited and viewed code).
    \item \textbf{S3r1-soft / S3r1-hard}: S3r0 plus one EM round with $n^\star=2$ or $1$.
    \item \textbf{S3r1-ctrl}: S3r0 plus the same extra training on the teacher's localization only. Soft against ctrl isolates the effect of the EM targets.
\end{itemize}

\paragraph*{Stage~3 as run.} Stage~3 uses three kinds of localizer and fixer. (1) \emph{Fold models}: we split the training tasks into two halves, A and B, and train one localizer and one fixer on each half (the fixers with the S2b recipe). (2) \emph{S3r0}: a localizer trained on all tasks with the teacher's localization (\stage{s3\_loc\_r0}). (3) \emph{The S2b fixer}, trained on all tasks. For every task, the localizer of the \emph{other} half proposes $K=8$ answers (temperature 0.8); we add the teacher's localization and a few simple variants, which gives 7.1 different candidates per task on average. The fixer of the other half then scores every candidate. One of the four scoring jobs stopped early, so 1,040 of the tasks have scores; the other tasks keep only the teacher's localization in Round~1. S3r1-soft and S3r1-hard are S3r0 fine-tuned on these weighted candidates; S3r1-ctrl is S3r0 fine-tuned on the teacher's localization again. The fold models only propose and score; at evaluation, every S3 localizer works with the S2b fixer.

\paragraph*{Evaluation.} We use the 78 development tasks whose reference fix passes its tests in our Kaggle environment (of 129). Each run uses the competition harness on 4$\times$L4 GPUs (vLLM, tensor parallel 4), 8 minutes per task, 36 tool calls for one agent and 44 for two, and 4 tasks in parallel. A task is solved when its hidden tests pass. Runs used two sampling settings: temperature 0.2-0.3 with an 8,192-token output limit, and temperature 0.7 with 6,656. We compare systems only within a setting, with the exact McNemar test~\citep{mcnemar1947} on the same tasks.

\section{Results}
\label{sec:results}

\subsection{Offline: the EM target helps the fixer as much, with less code}
\label{sec:signaltest}
Before training on the EM targets we ran an offline check on 1,016 training tasks. For each task, the other fold's fixer ran live from two inputs: the teacher's localization (T) or the highest-weight candidate (S). We then compared where it edited with the verified patch (\Cref{tab:signal}). The two inputs are equally useful: the first edit lands at a patched location in 43.1\% of attempts with S and 42.4\% with T ($p=0.74$). But S is 27.5\% smaller and causes 44\% fewer context overflows. The fixer also prefers a candidate other than the teacher's on 83\% of tasks, and on 86\% even without the size penalty; most often it prefers the localizer's own proposals (43\%) or the edited functions alone (26\%). Writing the exact verified patch is too rare at this model size (1\%) to compare.

\begin{table}[tbp]
\centering
\small
\caption{Offline signal test on 1,016 training tasks: the other fold's fixer starts from the teacher's localization (T) or the highest-weight candidate (S). Only-T / only-S count tasks where exactly one attempt succeeds; $p$ is the exact McNemar test.}
\label{tab:signal}
\begin{tabular}{lccrrc}
\toprule
Measure & T & S & Only T & Only S & $p$ \\
\midrule
Every hunk of $\ystar$ reproduced & 1.0\% & 1.3\% & 3 & 6 & 0.51 \\
At least one hunk reproduced & 5.3\% & 4.9\% & 24 & 20 & 0.65 \\
First edit at a patched location & 42.4\% & 43.1\% & 155 & 162 & 0.74 \\
Any edit & 55.3\% & 57.0\% & 168 & 185 & 0.39 \\
\midrule
Localization size (tokens) & 2,731 & 1,980 & \multicolumn{3}{l}{$-27.5\%$} \\
Context overflows & 146 & 82 & \multicolumn{3}{l}{$-44\%$} \\
\bottomrule
\end{tabular}
\end{table}

\subsection{End to end on the development set}
\label{sec:main}
\Cref{tab:main} shows all 13 runs. Three results stand out.

\begin{table}[tbp]
\centering
\small
\caption{All runs on the 78 development tasks (Gemma~4 E4B, 4$\times$L4, 8 minutes per task). \emph{Out of budget}: timeouts plus tasks that used all tool calls. \emph{Loops}: tasks with at least one output that reached the output limit. \emph{src/}: tasks where the agent used a \texttt{/workspace/src/} path that does not exist. \emph{No file}: tasks with five or more ``No such file'' errors. \emph{vs.\ base}: tasks solved only by the base / only by the system, with the McNemar $p$.}
\label{tab:main}
\resizebox{\linewidth}{!}{%
\begin{tabular}{lcccrrrrrrrl}
\toprule
System & Temp. & Output & Calls & \textbf{Solved} & Out of budget & Wrong patch & Median calls & Loops & src/ & No file & vs.\ base ($p$) \\
\midrule
Base & 0.2 & 8,192 & 36 & 8 & 2 & 61 & 11 & -- & -- & -- & -- \\
S1 & 0.2 & 8,192 & 36 & 10 & 54 & 14 & 30.5 & 13 & 6 & 4 & 2 / 4 (0.69) \\
S1 & 0.2 & 6,656 & 36 & \textbf{11} & 56 & 9 & 31.5 & 18 & 6 & 7 & 2 / 5 (0.45) \\
S2b & 0.3 & 8,192 & 44 & 6 & 54 & 14 & 32.5 & 28 & 34 & 27 & 4 / 2 (0.69) \\
S3r1-soft & 0.2 & 8,192 & 44 & 8 & 56 & 10 & 38.5 & 35 & 36 & 39 & 4 / 2 (0.69) \\
S3r1-hard & 0.2 & 8,192 & 44 & 4 & 60 & 10 & 42.5 & 29 & 34 & 33 & 5 / 1 (0.22) \\
\midrule
S1 & 0.6 & 6,656 & 32 & 8 & 24 & 40 & 32 & 10 & 6 & 2 & 5 / 5 (1.00) \\
S2a & 0.7 & 6,656 & 44 & 7 & 38 & 27 & 44 & 12 & 41 & 38 & 3 / 2 (1.00) \\
S2a-view & 0.7 & 6,656 & 44 & 4 & 37 & 25 & 44 & 13 & 43 & 44 & 7 / 3 (0.34) \\
S3r0 & 0.7 & 6,656 & 44 & 5 & 34 & 31 & 44 & 12 & 24 & 32 & 5 / 2 (0.45) \\
S3r1-ctrl & 0.7 & 6,656 & 44 & 6 & 32 & 26 & 44 & 16 & 28 & 54 & 6 / 2 (0.29) \\
S3r1-soft & 0.7 & 6,656 & 44 & 6 & 34 & 28 & 44 & 14 & 35 & 43 & 4 / 2 (0.69) \\
S3r1-soft & 0.7 & 6,656 & 32 & 4 & 34 & 31 & 32 & 15 & 34 & 44 & 5 / 1 (0.22) \\
\bottomrule
\end{tabular}}
\end{table}

\paragraph*{The EM targets move the agent in the right direction, but the gain in solved tasks is not significant.} Under identical settings (temperature 0.7, 44 calls), S3r1-soft solves 8 tasks and the matched control S3r1-ctrl solves 6; soft solves 4 tasks that ctrl does not and ctrl 2 that soft does not ($p=0.69$). The process measures agree with the offline test: soft reaches a patch on 53 tasks against 46, ends with no patch on 9 against 16, and has repeated ``No such file'' errors on 43 tasks against 54. Hard EM is not better than soft.

\paragraph*{The single agent is the most robust system.} S1 solves 8, 10 and 11 tasks in three runs. Under the base model's settings (temperature 0.2, 36 calls) it solves 10 and 11 against 8 for the base ($p=0.69$ and $0.45$). Every two-agent system solves 4-8 tasks.

\paragraph*{Differences of a few tasks are noise.} The same S1 model solves 8 to 11 tasks across runs; even its two runs at temperature 0.2 each solve 2-3 tasks that the other does not. Across all 13 runs, only 18 different tasks were ever solved. None of the comparisons in \cref{tab:main} reaches $p<0.05$.

\subsection{What limits a 4B agent here}
\label{sec:failures}
The trace of every run lets us see where the 8 minutes go. Four problems explain most failures.

\paragraph*{1. Trained agents explore, then run out of time.} The base model makes a median of 11 tool calls, finishes in about 2.5 minutes and submits a guess: 61 of 78 tasks end with a patch that fails the tests. Every trained system makes about three times as many calls (30-44) and makes far fewer wrong patches (9-14 at low temperature). But 54-60 of 78 tasks then run out of budget. The teacher needed a median of 15 calls within the same 8 minutes, on a fast API. On 4$\times$L4, with 4 tasks in parallel, the student generates about 50 tokens per second per task, so it imitates a pace it cannot keep. The scorer runs one task at a time, so our timeout counts are pessimistic, but the gap is large. 

\paragraph*{2. Low temperature causes repetition loops.} At temperature 0.2-0.3, 13--35 tasks per run contain an output that repeats one line until the output limit; each such output costs 130-150 seconds, more than a quarter of the budget. Two-agent systems loop about twice as often as S1 (28-35 tasks against 13-18). At temperature 0.6-0.7, loops drop to 10-16 tasks for every system. Only 16 of 222 looping outputs in the two-agent runs contain one of our fixed sentences, so the fixed text is not the main cause.

\paragraph*{3. Temperature trades timeouts for wrong patches.} Moving S1 from 0.2 to 0.6 cuts the out-of-budget tasks from 54-56 to 24, but wrong patches rise from 9-14 to 40, and solved tasks do not rise. The same holds for the two-agent systems (out of budget 54-60 $\to$ 32-38; wrong patches 10--14 $\to$ 25--31). Once the agent has time, the next limit is the quality of its edits.

\paragraph*{4. The trained localizers assume a code layout.} Every two-agent system uses paths under \texttt{/workspace/src/} that do not exist in 24-43 tasks, for example in fastapi and rich, and gets five or more ``No such file'' errors in 27-54 tasks. S1 does this in 6 and 2-7 tasks. The localizers keep the assumption even after listing the repository. The S3 localizers do it less than the S2 ones at the same settings (24-35 against 41-43 tasks), which fits the offline finding that the EM target is closer to what the fixer needs.

\section{Discussion}
\label{sec:discussion}

\paragraph*{Why the offline gain does not become more solved tasks.} The signal test and the EM scores measure whether the fixer is looking at the right code. At E4B scale, most tasks fail later: the agent runs out of time, loops, or writes a wrong edit (\Cref{sec:failures}). A better localizer cannot fix these. The EM round did improve what it can affect, reaching a patch and wasting fewer reads, which suggests the method is sound and the bottleneck is elsewhere for a 4B model.

\paragraph*{Lessons for others.}
\begin{itemize}[leftmargin=*]
    \item \textbf{Match the teacher's pace to the student's hardware.} A trajectory that fits in 8 minutes on a fast API does not fit on 4$\times$L4. Filter or shorten trajectories by the student's expected time, not the teacher's.
    \item \textbf{Leave room in the context window.} The harness asks for up to 16,384 output tokens with a 32K window, so any prompt above 16K tokens is rejected; in our first run this ended 4 of 5 tasks. An output limit of 6-8K avoids it. On this serving path the thinking budget was not enforced: outputs reached the full limit with a 1,536-token budget.
    \item \textbf{Run small evaluations more than once.} On 78 tasks the same model varies by three solved tasks; report paired tests and the set of solved tasks, not only the rate.
\end{itemize}

\section{Limitations}
\label{sec:limitations}

 We trained one model size (E4B) and ran one EM round that updated only the localizer. Each setting was run once per system, and the two groups of runs use different sampling settings, so we compare only within a group. We evaluated 4 tasks in parallel, which is slower per task than the scorer. The signal test measures where the fixer edits, not whether its patch passes. Candidate localizations were sampled after the teacher's exploration, not after the localizer's own.

\section{Conclusion}
\label{sec:conclusion}

We proposed \method, which trains a code localizer on what makes a verified patch likely under the fixer, using only weighted supervised fine-tuning. Offline, its target helps the fixer as much as the teacher's localization with 27.5\% less code; end to end on Gemma~4 E4B it moves the agent the right way (more patches, fewer wasted reads) but the gain in solved tasks is not significant. Our analysis of 13 runs shows that, for a 4B model under an 8-minute budget on 4$\times$L4, time, repetition loops and edit quality limit the agent more than localization does. We release the 1,629 verified trajectories, the data and training code, and all trained adapters. Larger models, further EM rounds and fixer updates are left to a follow-up paper.

\bibliography{main}

\clearpage
\newpage
\appendix

\section{Harness Facts that Shape Training}
\label{app:harness}

\begin{table}[H]
\centering
\small
\caption{Behaviour of the competition harness (swegemma 0.2.11, adk\_submission 0.2.13, adk\_eval\_core 0.1.0, google-adk 1.39.1), verified from the wheels and on Kaggle, and its consequence for training.}
\label{tab:harness}
\begin{tabular}{p{7.4cm}|p{7.4cm}}
\toprule
Fact & Consequence for training \\
\midrule
The whole trajectory renders as one model turn; tool calls and results are special-token spans inside it. & Training text is built from the exact messages the client sends, rendered with the model's own chat template. \\
\hline
Thinking is shown only for assistant messages with tool calls after the last user message. & Records are cut into segments at every new user message. \\ \hline
Visible text on a tool-call message renders after the tool responses. & Teacher narration on tool-call turns is folded into the thinking channel. \\ \hline
ADK appends ``You are an agent. Your internal name is \ldots'' to the system prompt. & Replay routes each request to the right agent's script by this line. \\ \hline
Counted tools: \texttt{run\_command}, \texttt{read\_file}, \texttt{edit\_file}, \texttt{write\_file}; not counted: \texttt{get\_status}, \texttt{submit\_patch}. & Budget messages are rewritten per counted call. \\ \hline
A budget warning is appended to a tool response when 10 or fewer calls remain (budget at least 20). & Re-generated with the harness's own function for the training budget. \\ \hline
The time budget is a hard timeout; afterwards the current diff is graded. & A cut-off run can still score if it edited early. \\ \hline
Compaction: interval 15, overlap 2, threshold 14,336 tokens, retention 5; the root agent writes the summary. & Contexts above about 14K tokens are summarized at inference; no compaction inside the localizer. \\ \hline
\texttt{read\_file}: at most 150 lines and 10,000 characters, cut mid-line at the character cap while \texttt{end\_line} reports the line cap. & Synthetic reads are chunked so every output is complete. \\
\texttt{run\_command}: stdout and stderr each cut at 5,000 characters; 300\,s timeout. & Used when rebuilding code from traces. \\ \hline
A sub-agent tool returns \texttt{\{"result": "<final non-thought text>"\}}. & Fixer records contain exactly this wrapper. \\ \hline
\texttt{skip\_summarization: true} ends the fixer's turn after the localizer call and triggers a nudge. & The localizer tool uses \texttt{skip\_summarization: false}. \\ \hline
LoRA modules: 258 on E4B (KV-shared layers have no k/v projections). & Adapters cover only the projections that exist. \\ \hline
E4B on 4$\times$L4: 298,464 tokens of KV cache. & Several full 32K sessions fit at once. \\
\bottomrule
\end{tabular}
\end{table}

\section{Training Settings and Budget Sweeps}
\label{app:settings}

\begin{table}[H]
\centering
\small
\caption{Choices of budget and length, and the share of trajectories (and tokens) each option keeps.}
\label{tab:budget}
\begin{tabular}{p{3.0cm}p{6.2cm}lp{3.0cm}}
\toprule
Setting & Options measured & Chosen & Reason \\
\midrule
Tool calls, single agent & 24 / 28 / 30 / 32 / 36 / 40 calls keep 78.0 / 85.0 / 86.8 / 89.2 / 92.5 / 94.8\% & 36 & keeps the hard tasks \\
Tool calls, two agents & -- & 44 & +8 for re-reading $z$ \\
Time budget & 5 or 8 minutes & 8 & the teacher's own limit \\
Localizer exploration & 12 calls keep 86.8\%, 15 keep 92.1\% & 15 & exploration p90 is 14 \\
Maximum length & trajectories / tokens kept: 24K 86.2 / 70.9\%, 28K 90.8 / 78.5\%, 32K 93.8 / 84.0\% & 32,768 & 24K drops 29\% of tokens \\
Compaction in records & replay with compaction vs.\ full trajectories & full & chosen; see limitations \\
\bottomrule
\end{tabular}
\end{table}

\begin{table}[H]
\centering
\small
\caption{Settings in use.}
\label{tab:settings}
\begin{tabular}{p{5.2cm}p{9.0cm}}
\toprule
Setting & Value \\
\midrule
Budget in records & single agent 36 calls / 8 min; two agents 44 calls / 8 min; 300 turns \\
Maximum sequence length & 32,768 \\
Localizer calls / locations / context budget & 15 / 8 / 6,000 tokens \\
Noisy fixer records without requests & 40\% \\
Candidates per task in Stage~3 ($K$) & 8 samples at temperature 0.8 from the other fold's localizer, plus $z^{\mathrm{teacher}}$ and perturbations \\
Target ESS $n^{\star}$ & 2 (soft), 1 (hard) \\
$\lambda_{\mathrm{ctx}}$, $\lambda_{\mathrm{req}}$ & $0.25\,\sigma$ per median-size $z$ and per request: 0.0097 per 1K tokens, 0.017 per request ($\sigma=0.068$) \\
Round-1 records & top $M=3$ candidates (weight $\ge 0.01$) + anchor $\beta=0.5$; total weight 1 per task \\
LoRA & rank 64, $\alpha$ 32, dropout 0; q, k, v, o, gate, up, down \\
Peak learning rate & $2\times10^{-4}$ (no sweep); $0.5\times$ for warm starts (Round~1 and control) \\
Optimizer and schedule & AdamW (0.9, 0.999), no weight decay, clip 1.0; 10 warm-up steps, cosine to 10\% \\
Batch, epochs & 16 rows per optimizer step (4 GPUs $\times$ 1 row $\times$ 4 accumulation steps), no packing, length grouping; 3 epochs for localizers and S1, 1 epoch for fixers trained on gold + noisy records and for Round~1 \\
Hardware & 4$\times$A100 80\,GB (training, Stage~3 generation and scoring); Kaggle for evaluation \\
Model & Gemma~4 E4B only \\
Validation split & 3\% of tasks \\
Seed & 1234 \\
\bottomrule
\end{tabular}
\end{table}

\section{Record Filters and Drop Counts}
\label{app:filters}

\begin{table}[H]
\centering
\small
\caption{Filters applied when building records.}
\label{tab:filters}
\begin{tabular}{p{4.4cm}p{2.2cm}p{8.0cm}}
\toprule
Filter & Applies to & Meaning \\
\midrule
over budget & both & more counted calls than the budget (two agents: localizer and fixer together) \\
too long & both & single agent: any record above 32,768 tokens drops the trajectory; two agents: that record alone \\
localizer over calls & two agents & exploration longer than 15 calls \\
impossible input & two agents & $z$ misses an edited location and no request recovers it \\
unseen code & two agents & code rebuilt from the trace, and a read needs a line the teacher never printed \\
missing code & two agents & no code for an edited file \\
dev repository, not in solved set, replay error & both & development-set repository; dropped by the cleaning audit; trace cannot be loaded or replayed \\
\bottomrule
\end{tabular}
\end{table}

Two-agent drops, counted per variant (a task can appear several times): localizer exploration over 15 calls, 51-54 per variant; over the 44-call shared budget, 15-35 per variant; view-only $z$ missing an edited location with no way to recover it, 80; code a read needs but the teacher never printed, 1-2 per variant; fixer records over 32K tokens, 11 gold and 11 noisy. Tasks without code for an edited file: 3; without any edit: 3; with the first edit mixed into an exploration turn: 1.

\section{Data Funnel}
\label{app:funnel}

\begin{table}[H]
\centering
\small
\caption{From collected tasks to training records.}
\label{tab:funnel}
\begin{tabular}{lr}
\toprule
Step & Count \\
\midrule
Tasks collected from seven datasets & $\approx$10K \\
Passed ingest checks (patches apply at the base commit) & 9,966 \\
Ranked candidates under source and repository caps & 7,772 \\
Valid among the first 4,500 ranked (tests fail before, pass after the gold patch) & 3,894 \\
Tasks attempted by the teacher (attempts) & 2,959 (3,068) \\
Tasks solved & 1,629 (55\%) \\
Solved tasks accepted under the caps & 1,577 \\
Trajectories with single-agent records & 1,384 \\
Tasks with two-agent records & $\approx$1,300 \\
Unsolved attempts kept for Stage~4 (with code) & 1,338 (1,310) \\
\bottomrule
\end{tabular}
\end{table}

\section{Issues Found During Development}
\label{app:bugs}

\begin{table}[H]
\centering
\small
\caption{Problems found while building the pipeline, their effect if missed, and the fix.}
\label{tab:bugs}
\begin{tabular}{p{4.6cm}p{4.6cm}p{5.6cm}}
\toprule
Problem & Effect if missed & Fix \\
\midrule
Localizer prompt said 12 calls; records keep up to 15 & training on examples that break their own instruction & prompt says 15; a test ties prompt numbers to the configuration; records rebuilt \\
No base-commit code for solved tasks & 622 of 1,384 tasks lost from two-agent data & fetch needed files with plain git, checked against the trace \\
SWE-smith commits not served by GitHub (300 tasks) & no code for SWE-smith tasks & instance branch of the SWE-smith mirror \\
Public vLLM 0.19.1 lacks LoRA for Gemma~4's model class & parity or evaluation differs from the scorer & install the competition wheel; build check \\
DDP broadcasts buffers by position; rotary buffer order depends on the hash seed & all GPUs but the first train with wrong positions & sort buffers by name before wrapping; regression test \\
40 single-agent traces failed replay with a server error & about 3\% of Stage~1 data lost; cause hidden & replay re-raises the underlying error; cause open \\
12 audit-dropped traces still produced records & training on rejected trajectories & builder skips traces not in the solved set \\
5-minute budget replayed on the teacher's 8-minute clock & teaches the model to ignore the clock & time scale = budget / teacher minutes (1.0 at 8 minutes) \\
\texttt{skip\_summarization: true} on the localizer tool & broken two-agent loop at inference & set to false; a test checks one user message per fixer record \\
\texttt{read\_file} cuts mid-line at 10,000 characters & synthetic reads that differ from the harness & reads chunked to 150 lines and 10,000 characters \\
End-of-epoch validation assumed one validation file & runs with two (S2b and fold fixers) stopped after saving the adapter, before export & per-file sources in validation; the saved adapter exported afterwards \\
vLLM does not reserve memory for prompt log-probabilities & out of memory on the first long scoring sequence & scoring at 55\% GPU memory utilization \\
Offline rollouts without compaction & server rejects prompts above about 16K tokens; more often for larger $z$ & such attempts counted as stopped, not excluded \\
One scoring shard stopped early & 1,040 instead of about 1,240 tasks with weights & unscored tasks keep $z^{\mathrm{teacher}}$ in Round~1 \\
Template drops earlier thoughts after a user message & training text differs from inference text & segments at user messages; narration folded into thinking \\
Output limit of 16,384 tokens with a 32,768-token window & every prompt above 16K tokens is rejected; 4 of the first 5 development tasks ended this way & output limit of 6,656--8,192 tokens \\
Low sampling temperature (0.2--0.3) & repetition loops in 13--35 of 78 tasks, each costing 130--150\,s & temperature 0.6--0.7 \\
\bottomrule
\end{tabular}
\end{table}

\section{Reproduction}
\label{app:repro}

Every step runs through one command-line tool; each command takes a configuration file and overrides, and writes its resolved configuration, git commit and package versions next to its output. The code is split into two repositories:
\begin{itemize}[leftmargin=*]
    \item \textbf{Data} (\url{https://github.com/khalil-Hennara/codeagent-validate}): task ingest and validation, teacher runs and grading, recall cleaning, trace statistics and budget sweeps, base-commit code fetch, and single-agent and two-agent record building in parallel shards, with merge, schema check and the task-level validation split.
    \item \textbf{Training} (\url{https://github.com/misraj-ai/locfix}): dequantization once per model; a GPU smoke test (dequantize, parity, overfit, export, parity, generation); a dry-run training plan; training; and a parity check of the exported adapter before any evaluation. Stage~3 adds the fold split (\texttt{em.fold-split}), candidate generation (\texttt{em.candidates}), scoring sequences (\texttt{score.prepare}), scoring (\texttt{score.run}), weights (\texttt{score.weights}), the offline signal test (\texttt{em.signal-test}) and the Round-1 records (\texttt{em.r1-records}); \texttt{submit.package} builds and checks the submissions.
\end{itemize}
Versions: vLLM 0.19.1 (competition build), transformers 5.13.1, peft 0.21.2, accelerate 1.15.0, compressed-tensors 0.15.0.1, google-adk 1.39.1, litellm 1.106.3, Python 3.12 (training).

\end{document}